\documentclass{article} 
\usepackage[preprint]{colm2026_conference}
\usepackage{microtype}
\usepackage{hyperref}
\usepackage{url}
\usepackage{booktabs}

\usepackage[LGR,T1]{fontenc}  
\usepackage[greek,english]{babel}
\usepackage[utf8]{inputenc}
\usepackage{cjhebrew}

\usepackage{microtype}

\usepackage{inconsolata}
\usepackage[textsize=scriptsize,textwidth=2cm]{todonotes}

\usepackage{graphicx}
\usepackage{amsmath}
\usepackage{amsthm}
\usepackage{amsfonts}
\usepackage{amssymb}
\usepackage{tikz}
\usepackage{cleveref}
\usepackage{booktabs}
\usepackage{bbm}
\usepackage{mathtools}
\usepackage{multirow}
\usepackage{color,soul}
\usepackage{float} 
\usepackage{makecell}
\usepackage{array}
\newcolumntype{P}[1]{>{\raggedright\arraybackslash}p{#1}}
\usepackage[dvipsnames]{xcolor}
\usepackage{wrapfig}

\usepackage{titlesec}
\titlespacing*{\paragraph}{0ex}{0.5ex}{1ex}

\usepackage{enumitem}
\setlist[enumerate,itemize]{topsep=0pt,itemsep=0pt,leftmargin=18pt}

\usepackage{soul}

\usepackage{subcaption}
\usepackage{adjustbox}
\usepackage{examples-slim}

\theoremstyle{definition}

\DeclareMathSymbol{\shortminus}{\mathbin}{AMSa}{"39}
\usepackage{makecell} 

\crefformat{section}{\S#2#1#3}
\crefformat{subsection}{\S#2#1#3}
\crefformat{subsubsection}{\S#2#1#3}
\crefformat{paragraph}{\P#2#1#3}
\crefformat{subparagraph}{\P#2#1#3}
\crefmultiformat{section}{\S#2#1#3}{ and~\S#2#1#3}{, \S#2#1#3}{, and~\S#2#1#3}
\crefmultiformat{subsection}{\S#2#1#3}{ and~\S#2#1#3}{, \S#2#1#3}{, and~\S#2#1#3}
\crefmultiformat{subsubsection}{\S#2#1#3}{ and~\S#2#1#3}{, \S#2#1#3}{, and~\S#2#1#3}
\crefmultiformat{paragraph}{\P\P#2#1#3}{ and~#2#1#3}{, #2#1#3}{, and~#2#1#3}
\crefmultiformat{subparagraph}{\P\P#2#1#3}{ and~#2#1#3}{, #2#1#3}{, and~#2#1#3}
\crefrangeformat{section}{\mbox{\S\S#3#1#4--#5#2#6}}
\crefrangeformat{subsection}{\mbox{\S\S#3#1#4--#5#2#6}}
\crefrangeformat{subsubsection}{\mbox{\S\S#3#1#4--#5#2#6}}
\crefrangeformat{paragraph}{\mbox{\P\P#3#1#4--#5#2#6}}
\crefrangeformat{subparagraph}{\mbox{\P\P#3#1#4--#5#2#6}}
\crefname{part}{Part}{Parts}
\Crefname{part}{Part}{Parts}
\crefname{chapter}{Ch.}{Ch.}
\Crefname{chapter}{Ch.}{Ch.}
\crefname{footnote}{Fn.}{Fn.}
\Crefname{footnote}{Fn.}{Fn.}
\crefname{figure}{Figure}{Figures}
\crefname{table}{Table}{Tables}
\crefname{subfigure}{Figure}{Figures}
\Crefname{subfigure}{Figure}{Figures}
\crefname{appsec}{Appendix}{Appendices}
\Crefname{appsec}{Appendix}{Appendices}
\crefname{algocf}{Algorithm}{Algorithms}
\Crefname{algocf}{Algorithm}{Algorithms}
\crefname{xnumi}{ex.}{exs.}
\Crefname{xnumi}{Ex.}{Exs.}
\crefname{xnumii}{ex.}{exs.}
\Crefname{xnumii}{Ex.}{Exs.}

\usepackage{lineno}

\definecolor{darkblue}{rgb}{0, 0, 0.5}
\hypersetup{colorlinks=true, citecolor=darkblue, linkcolor=darkblue, urlcolor=darkblue}

\title{Causal Drawbridges: Characterizing Gradient Blocking of Syntactic Islands in Transformer LMs}

\author{Sasha Boguraev \quad \quad Kyle Mahowald \\
        The University of Texas at Austin\\
        \texttt{\{sasha.boguraev,kyle\}@utexas.edu} 
        }

\begin{document}

\ifcolmsubmission
\linenumbers
\fi

\maketitle

\begin{abstract}
We show how causal interventions in Transformer models provide insights into English syntax by focusing on a long-standing challenge for syntactic theory: syntactic islands.
Extraction from coordinated verb phrases is often degraded, yet acceptability varies gradiently with lexical content (e.g., ``I know what he hates art and loves'' vs. ``I know what he looked down and saw''). We show that modern Transformer language models replicate human judgments across this gradient. Using causal interventions that isolate functionally relevant subspaces in Transformer blocks, attention modules, and MLPs, we demonstrate that extraction from coordination islands engages the same filler–gap mechanisms as canonical \textit{wh}-dependencies, but that these mechanisms are selectively blocked to varying degrees. By projecting a large corpus of unrelated text onto these causally identified subspaces, we derive a novel linguistic hypothesis: the conjunction ``and'' is represented differently in extractable versus non-extractable constructions, corresponding to expressions encoding relational dependencies versus purely conjunctive uses. These results illustrate how mechanistic interpretability can inform syntax, generating new hypotheses about linguistic representation and processing.
\end{abstract}

\section{Introduction}
Our aim in this work is to study how, mechanistically, Language Models (LMs) handle \textit{syntactic islands}, as a paradigmatic case of a complex and gradient linguistic phenomenon that is nonetheless learned by these models. 
In addition to giving empirical evidence for fine-grained similarity in how LMs and humans process the construction, our main goal is to uncover the mechanism by which these phenomena are handled in LMs. We do so using causal intervention techniques to tell the story of how these phenomena are handled.

An enduring question in human language research is how human learners come to master long-distance filler-gap constructions like (\ref{ex:gap}). Contrast this with the gapless example in (\ref{ex:fg_nogap}). (\ref{ex:gap}) is considered a filler-gap construction since there is a gap (represented by the blank) that is licensed by (and in a sense ``filled by'') the \textit{wh}- word ``what''. 

\noindent
\begin{minipage}[t]{0.475\linewidth}
\begin{examples}
  \item I know [what]$_t$ she smiled at \_\_$_t$.\label{ex:gap}
\end{examples}
\end{minipage}%
\hfill
\begin{minipage}[t]{0.475\linewidth}
\begin{examples}
  \item I know [that] she smiled at it.\label{ex:fg_nogap}
\end{examples}
\end{minipage}

Sometimes, though, this filler-gap mechanism is blocked, leaving the sentence unacceptable (or at least questionably acceptable). Consider:

\begin{examples}
    \item \textbf{?} I know [what]$_t$ she hates art and loves \_\_$_t$.
    \label{ex:vp-island}
\end{examples}

\noindent This has a semantic interpretation, but syntactically sounds off.
This construction is called a syntactic island since the gap is ``stranded'' \citep{ross1967constraints}. 
The linguist Cedric Boeckx writes: ``Most modern grammarians would agree with me if I said that island effects are perhaps the most important empirical finding in modern theoretical linguistics'' \citep{boeckx2012syntactic}.
\begin{figure}
    \centering
    \includegraphics[width=\linewidth]{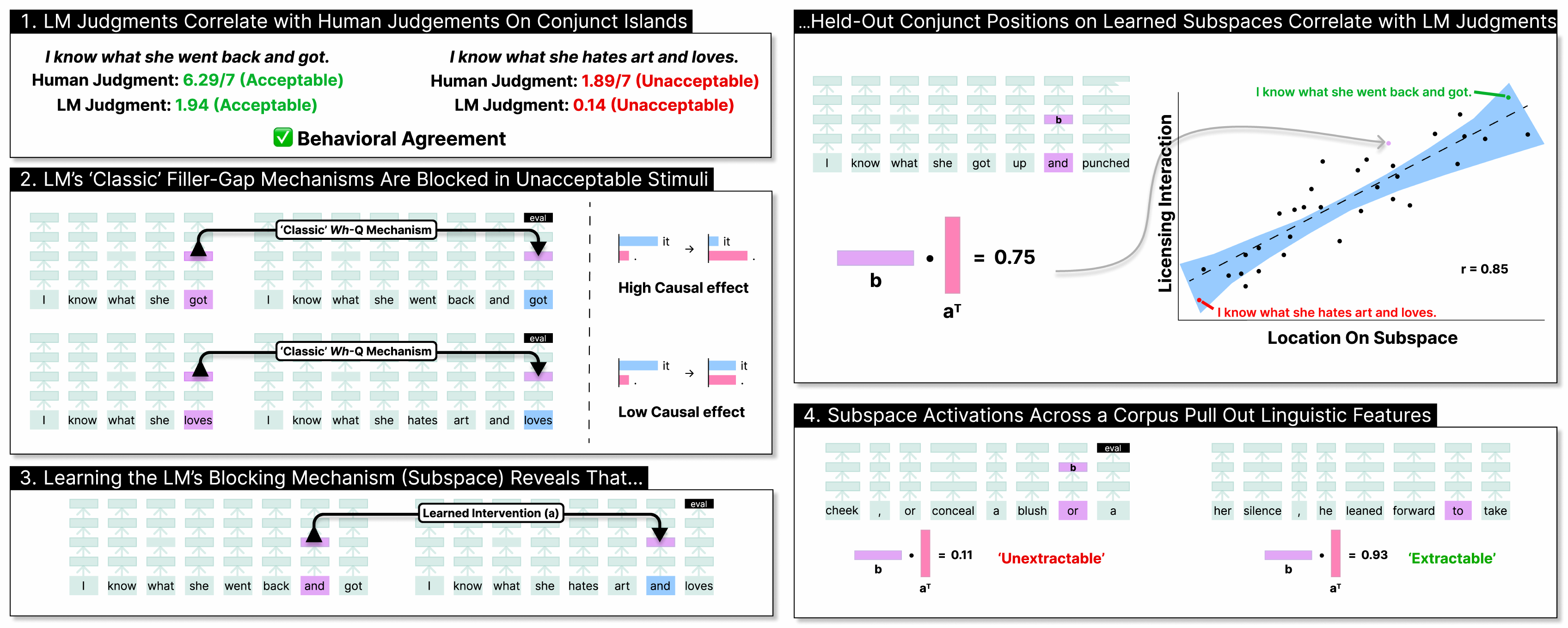}
    \caption{\textbf{1.} LM judgments of gradiently acceptable conjuncts correlate with human judgments. \textbf{2.} These constructions rely on a more general-purpose filler-gap mechanism, blocked in the island case. \textbf{3.} We identify the relevant blocking subspaces, showing they correlate with model acceptability, and \textbf{4.} pick out linguistically meaningful structures in a corpus.}
    \label{fig:placeholder}
    \vspace{-1em}
\end{figure}
There are several reasons for all the interest.
We posit that the reasons that make islands linguistically interesting also make them interesting for interpretability more generally.

First, linguistic theory suggests that filler-gap dependencies require rich abstract structure, to keep track of long-distance dependencies. To know how to end (\ref{ex:gap}), you have to keep track that there was a gap-licensing \textit{wh}-element (``what'').
Second, the \textit{blocking} of the filler-gap mechanism, as in the island case, requires learning what constructions block it. Here, it's the verb-phrase (VP) conjunct ``hates art and loves \_\_\_'' that forms the island.

Third, depending on the content of the VP conjunct, the filler-gap mechanism is not actually blocked. People typically accept (\ref{ex:extract}), suggesting that the strength of the island is dependent on the lexical content of the conjunct -- making these constructions a case where high-level syntactic generalization depends on lower-level lexical information.

{\exampleindent=3em
\begin{examples}
    \item I know [what]$_t$ she looked down and saw \_\_$_t$.
    \label{ex:extract}
\end{examples}
}

Fourth, filler-gaps are rare, and it's claimed children receive little positive evidence that syntactic islands are disallowed. Yet all adult English speakers learn as much. Filler-gaps are thus taken as examples \textit{par excellence} of the Poverty of Stimulus problem \citep{pearl2022poverty}.

As a result of this state of affairs, Boeckx writes: ``The sophisticated analyses [...] put forth in order to capture island effects offer perhaps the strongest case for a rich, abstract, domain-specific mental module for language'' \citep{boeckx2012syntactic}.

Given that, it would certainly be interesting if we could study how a fairly general-purpose learner trained on a finite amount of data processed these constructions.
Our results suggest that, for the specific case of coordination islands illustrated here, modern LMs provide such a system.
First, we show that they are gradiently sensitive, in human-like ways, to coordination islands both in the general case and in ways that vary with the lexical items.
Second, we use mechanistic interpretability techniques, specifically causal interventions, to show extraction from VP conjuncts is handled by classical filler-gap mechanisms which gets blocked in the island case. 
Third, we show we can identify the ``drawbridge'' mechanism necessary for stranding and unstranding syntactic islands.
Finally, we use the identified subspaces to find other similar activations in a larger corpus of text. 
We use this method to motivate the hypothesis that the conjunction in extractable and non-extractable conjuncts have subtly lexically distinct representations---a claim we argue could be tested in humans.

\section{Background}

\subsection{Syntactic Island Constraints}

Our focus is on islands that arise from the Coordinate-Structure-Constraint (CSC), i.e., `do not extract from a single conjunct of a coordinated structure' \citep{ross1967constraints}, as in (\ref{ex:vp-island}).

While Ross admits violations to various island constraints, the broad class of island effects are often described as universal constraints on language \citep{ross1967constraints, chomsky1977wh}. 
From this observation and analyses of the linguistic input, a Poverty of the Stimulus argument follows, suggesting that if a learner only encounters positive evidence (`acceptable' sentences) throughout development, they would be unable to learn these constraints without an innate linguistic faculty \citep{chomsky1973conditions, phillips2013nature,pearl2022poverty}.

Yet there is also a large body of evidence arguing for gradience in these effects and/or sensitivity to non-syntactic factors \citep[see, e.g.,][i.a.]{kluender1991cognitive, ambridge2008island,  abeille2020extraction, liu2022verb, liu2022structural, namboodiripad2022backgroundedness, cuneo2023discourse, fergus2025islands}, such as cognitive processing \citep{kluender1993subjacency, hofmeister2013islands} or discourse effects \citep{kehler2002coherence,goldberg2006constructions, goldberg2011backgrounded, abeille2020extraction}.

Our work builds upon that of \citet{fergus2025islands}, whose materials and human ratings we use.
They showed that while the CSC contests that one cannot extract the argument of a single conjunct within a verb phrase (correctly predicting (\ref{ex:vp-island}) as ungrammatical), across a set of 46 such coordinated verb phrases, human acceptability judgments are gradient.
They suggest the gradience is explained by the degree to which the coordinated VP represents a single complex state that matters, but also note the potential effect of the first verb's transitivity -- many of the more acceptable variants have intransitive verbs as the first verb. 

\subsection{LMs and the Filler-Gap Task}

As interest grew in whether LMs can process complex syntax, a cottage industry emerged assessing whether they could handle long-distance constructions \citep{linzen2016assessing} and, in particular, filler-gaps and island constraints \citep[e.g.,][]{wilcox2018rnn,futrell2019neural,warstadt2020blimp,kobzeva2023neural, wilcox2023using}.
While success was mixed on early LMs, today's LMs successfully handle these constructions \citep[c.f.][]{lan2024bridging}.

As a result, they have also been the locus for work on neural interpretability, with a focus on \textit{how} they do so.
In particular, there has been a focus on whether the filler-gap mechanism is shared across linguistically related long-distance constructions \citep{howitt-etal-2024-generalizations,lan2024bridging,boguraev-etal-2025-causal,prasad-etal-2019-using,bhattacharya-van-schijndel-2020-filler}.

Unlike this prior work, our goal is not to study how coordination islands relate to other constructions, but to zoom in on the construction with as much granularity as we can.
We want to understand how the filler-gap mechanism operates in these constructions, how it is blocked in the island setting, and how it is unblocked in sentences that look superficially like islands but are not, as in \citet{fergus2025islands}.
To do that, we separately consider the Output of Transformer Blocks and Attention Modules, as well as MLP Activations -- further examining how these hidden features correspond to linguistic features.
Our goal in doing so is in part to understand how the particulars of the Transformer architecture are relevant in this task \citep[as in][\textit{i.a.}]{wang2023interpretability,elhage2021mathematical}.

\section{General Methods}

\subsection{Data}

We focus on verb-phrase conjuncts -- and corresponding human-ratings -- collected by \citet{fergus2025islands}. Their stimuli span 46 conjuncts, with Likert scale ratings from 1-7. 
Ratings are sourced from 442 participants with about twenty individuals judging each conjunct. 

These stimuli are in the form of matrix \textit{wh}-questions, making corresponding minimal pairs difficult to generate. As such, we transform each conjunct into the corresponding embedded \textit{wh}-questions.
We adapt the resulting data to sentential templates à la \citet{arora2024causalgymbenchmarkingcausalinterpretability} and \citet{boguraev-etal-2025-causal} for ease of sampling large numbers of unique minimally-paired data (see \Cref{app:conversion} for details).
We filter out all such stimuli in which the extracted \textit{wh}-item is not \textit{what} for consistency and those with continuations after the extracted object due to inherent limitations of autoregressive LMs. This leaves us with 33 minimally-paired templates. 

\subsection{Behavioral Analysis}\label{sec:behavior-setup}
\leavevmode
\begin{wrapfigure}{r}{0.3\linewidth}
    \vspace{-1em}
    \centering
    \includegraphics[width=\linewidth]{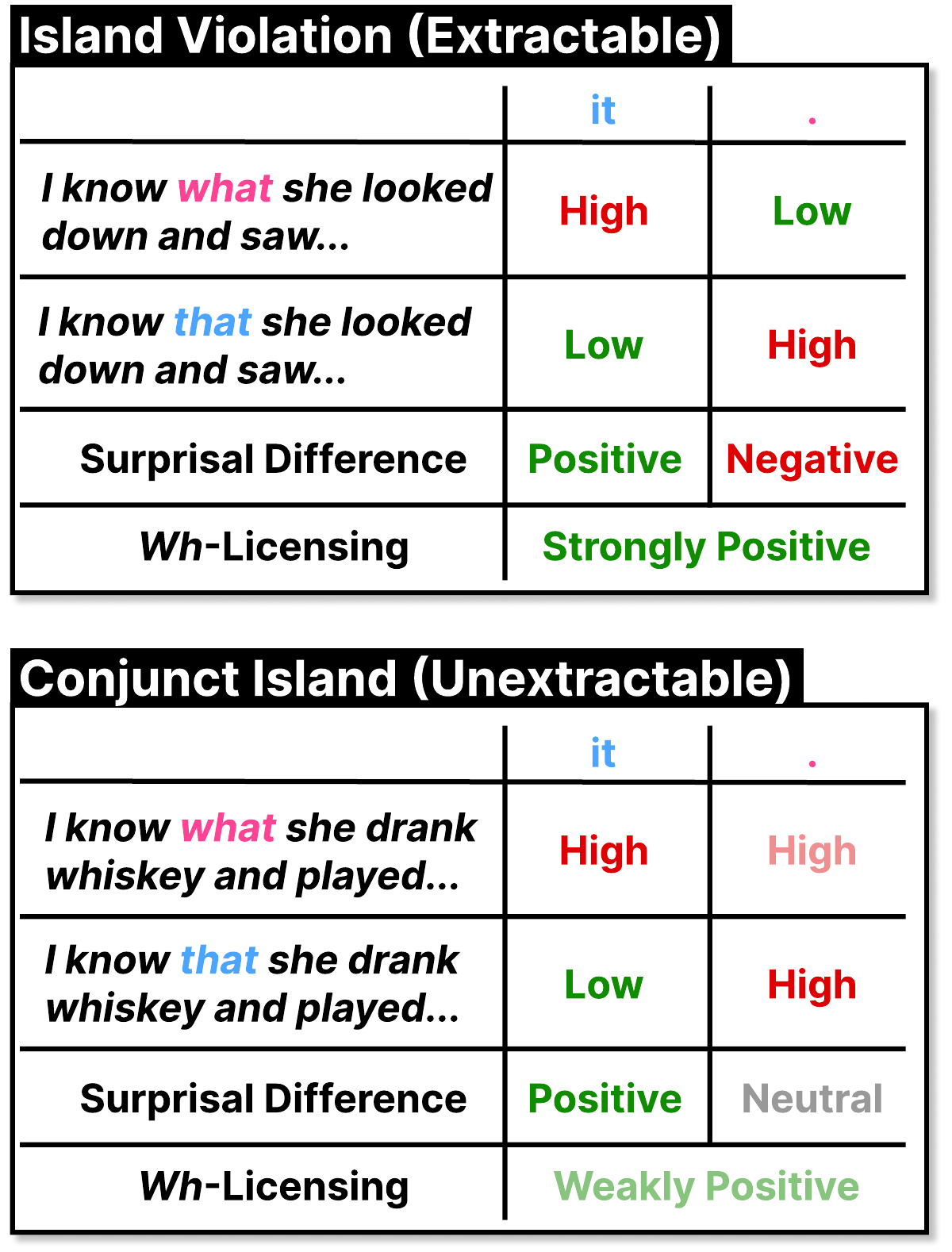}
    \vspace{-1em}
    \caption{Exemplar \textit{wh}-licensing calculations for extractable (top) and unextractable (bottom) conjuncts respectively.}
    \label{fig:wh-schema}
    \vspace{-1em}
\end{wrapfigure}
To measure how robustly an LM represents extraction for given stimuli, we follow the methodology of \citet{wilcox2018rnn}.
Specifically, we get minimal sentence pairs: $wh$, which contains a \textit{wh}-licensor, and $th$, which does not.
They have corresponding labels $l_{wh}$ (gap) and $l_{th}$ (no gap). We then compute the model’s surprisal, $S$, for each label given each sentence.

This 2$\times$2 setup allows us to calculate the gap-less interaction: the surprisal difference of a gap-less continuation for a sentence with a \textit{wh}-licensor, $S(l_{th}|wh)$, and without a licensor, $S(l_{th}|th)$. It also allows us to calculate the gap interaction: the surprisal difference for a gap continuation when comparing sentences with a \textit{wh}-licensor, $S(l_{wh}|wh)$, to those without one, $S(l_{wh}|th)$. If an LM robustly represents a long-distance dependency, the gap-less interaction, $S(l_{th}|wh) - S(l_{th}|th)$, should be positive, and the gap interaction, $S(l_{wh}|wh) - S(l_{wh}|th)$, negative. The overall \textit{wh}-licensing term is then the difference between these effects: $(S(l_{th}|wh) - S(l_{th}|th)) - (S(l_{wh}|wh) - S(l_{wh}|th))$. Higher values of this term indicate more salient representations of the long-distance dependency in the model. A schematic of this method is in \Cref{fig:wh-schema}.\footnote{Code for replicating experiments is at \url{https://github.com/SashaBoguraev/causal-drawbridges}.}

\subsection{Interpretability Analysis}\label{sec:interp-setup}

\paragraph{Distributed Alignment Search}

To localize the internal mechanisms utilized by LMs to process these conjuncts, we use Distributed Alignment Search \citep[DAS;][]{geiger2024finding}, consistent with prior work on syntactic interpretability \citep{arora2024causalgymbenchmarkingcausalinterpretability,boguraev-etal-2025-causal}. DAS is a supervised interpretability method which finds $d$-dimensional subspaces in an LM causally responsible for specified behavior. In particular, DAS acts on a base value $b\in \mathbb{R}^n$ -- the tensor at a given internal site when the model processes an input (e.g., Example \ref{ex:gap}) -- and corresponding source value $s\in\mathbb{R}^n$ -- the tensor at the same site when the model processes a minimally-paired input (e.g., Example \ref{ex:fg_nogap}). The intervention performed by DAS is 

\begin{equation*}
    \textbf{b} + (\textbf{sa}^\top - \textbf{ba}^\top)\textbf{a}
\end{equation*}

\noindent where $a\in\mathbb{R}^{n\times d}$ is a rotation learned through gradient-descent on the cross-entropy loss of the LM's prediction under intervention. Intuitively, DAS aims to find subspaces on which intervening from $s$ to $b$ 
maximizes the corresponding prediction, $l_s$. We learn 1D interventions, localizing features in three sites: (1) the output of Transformer blocks, (2) the output of attention-heads before head-mixing \citep[i.e., an LM's OV-circuit outputs; ][]{elhage2021mathematical} and (3) MLP activations. We choose these sites to probe how LMs' different sub-components process these syntactic phenomena in addition to their joint processing.

\paragraph{Training}

We train DAS at each position of our minimal pairs and layer of our LMs. For every intervention, we also learn a control intervention trained on the same minimal pairs but flipped labels. By comparing efficacies, we can evaluate the degree to which our interventions represent the LMs true mechanisms, as opposed to merely the separability of the labels. We train 5 seeds for each intervention. 
For more training details see \cref{app:train-eval-detail}.

\paragraph{Evaluation}

We evaluate interventions with the \textbf{\textsc{Odds}} metric introduced by \citet{arora2024causalgymbenchmarkingcausalinterpretability}. This metric measures the increase in probability of the intervened label after an intervention, relative to the decrease in probability of the expected label. Higher \textbf{\textsc{Odds}} indicate higher causal efficacy. We report the average $\Delta$\textbf{\textsc{Odds}}, that is, the difference between our critical condition \textbf{\textsc{Odds}} and the corresponding control condition, averaged across seeds.

\section{Exp. 1: LMs Agree with Human Judgments of Coordination Islands}\label{sec:experiment-one}

We first examine whether LMs share human gradient acceptability judgments across filler-gap sentences with conjoined verb phrases. 

\begin{wrapfigure}{r}{0.425\linewidth}
    \vspace{-1em}
    \centering
    \begin{subfigure}{\linewidth}
        \centering
        \includegraphics[width=\linewidth]{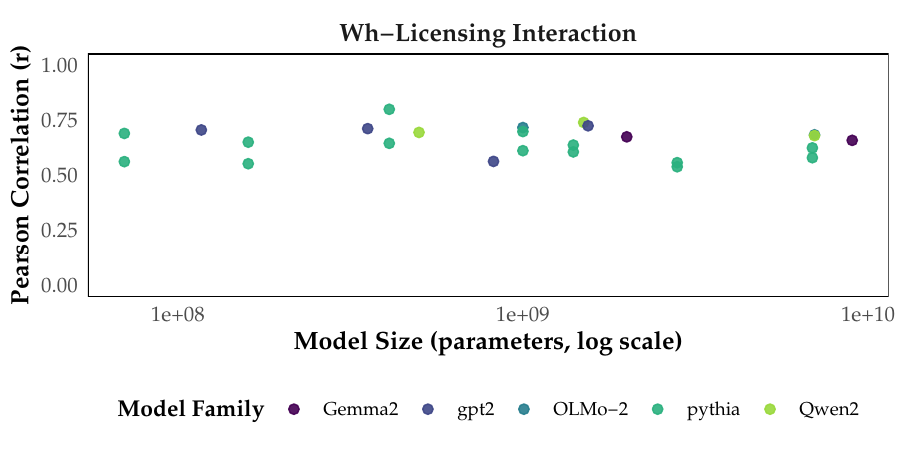}
        \caption{Pearson $r$ correlations between LM \textit{wh}-licensing interaction and human acceptability judgments, shown as a function of model size. Full scatter plots are in \cref{app:human-corr}.}
        \label{fig:behavior-corr}
    \end{subfigure}
    \begin{subfigure}{\linewidth}
        \centering
        \includegraphics[width=\linewidth]{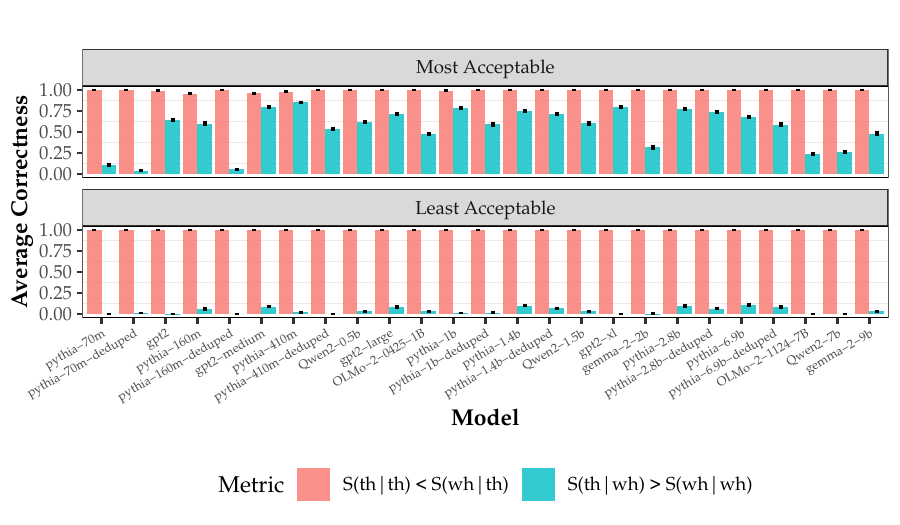}
        \caption{LMs' predicting of the correct label given a stimulus for the four most- and least-extractable conjuncts (measured by human acceptability), ordered left-to-right by size. All models perform well in the gap-less case (red), while for the gap cases, performance is gradiently related to extractability in the expected direction (green).}
        \label{fig:behavior-avg}
    \end{subfigure}
    \vspace{-1.5em}
    \caption{LM behavioral metrics.}
    \label{fig:behavior-combined}
    \vspace{-3em}
\end{wrapfigure}

\paragraph{Methods}
We sample 400 minimal pairs of each conjunct and calculate the mean \textit{wh}-interaction across pairs as described in \cref{sec:behavior-setup}. 
We then measure the correlation (Pearson $r$-value) between human judgments and an LM's mean \textit{wh}-interaction.
We test 25 LMs across 5 model families: \texttt{OLMo2} \citep{olmo20252olmo2furious}, \texttt{Qwen2} \citep{yang2024qwen2technicalreport}, \texttt{Gemma 2} \citep{gemmateam2024gemma2improvingopen}, \texttt{gpt2} \citep{radford2019language} and \texttt{pythia} \citep{biderman2023pythiasuiteanalyzinglarge}, testing a host of different sizes from 70 million to 9 billion parameters.

\paragraph{Results} Figure \ref{fig:behavior-corr} shows strong correlation between LM \textit{wh}-licensing and human acceptability judgments (between 0.54 and 0.80). 
Figure \ref{fig:behavior-avg} further plots the proportion that an LM's surprisal for a given stimuli's correct label is less than the mismatched label, averaged across the four most and least acceptable conjuncts (as judged by people).

Our \textit{wh}-licensing results suggest LMs broadly agree with human judgments. That is, generally, the more acceptable an extraction is deemed by humans, the more saliently LMs represent the long-distance dependency. This is reinforced by our binary comparison results. LMs consistently judge that for a $th$ stimuli, $l_{th}$ is more probable than $l_{wh}$, whether or not the minimally-paired $wh$ stimuli is acceptable or not (i.e., the red bars are high in both facets). Further, most LMs correctly display a preference for extraction in the extractable $wh$ case, and much less so in the non-extractable case. 

\paragraph{Discussion} 
In the next sections, we turn our eye towards developing a mechanistic understanding of exactly how LMs come to these judgments. We focus our analysis on three LMs with strong behavioral correlation: \texttt{gpt2}, \texttt{pythia-410m-deduped}, and \texttt{Qwen2-0.5b}.

\begin{figure*}[t]
    \centering

    \begin{subfigure}[t]{0.9\textwidth}
        \centering
        \includegraphics[width=\linewidth]{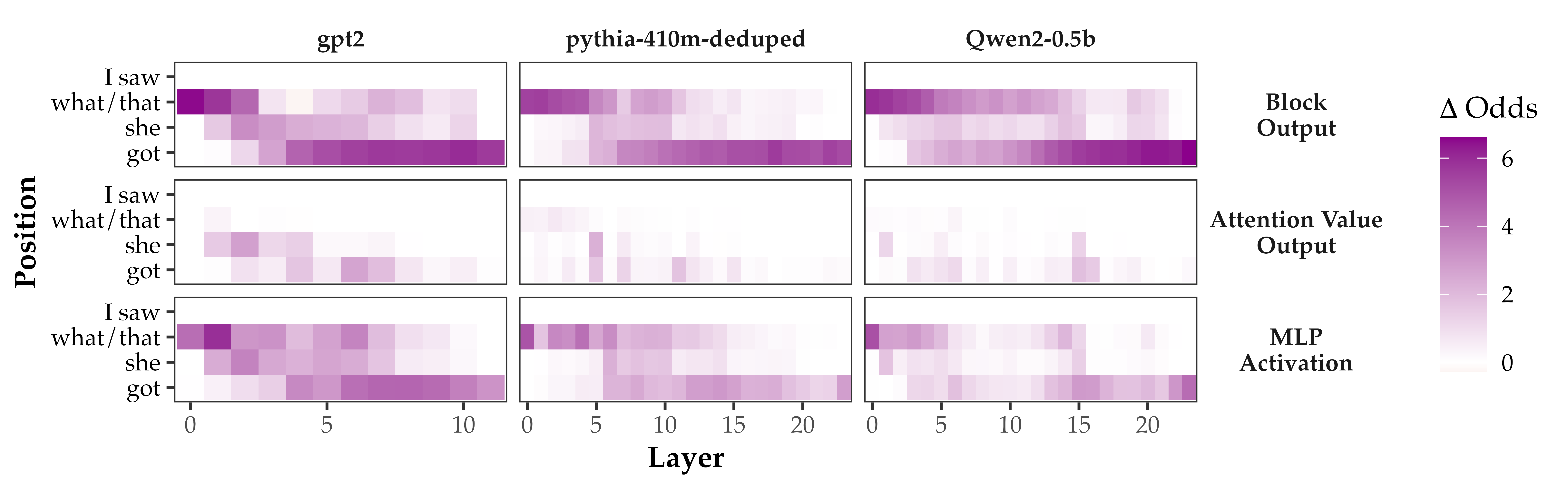}
        \caption{\textbf{\textsc{Odds}} for learned embedded \textit{wh}-question interventions evaluated on a held-out test set. Strong causal efficacy suggests DAS has robustly recovered the filler-gap mechanism.}
        \label{fig:generalization-middle}
    \end{subfigure}
    \vspace{-.55em}
    \begin{subfigure}[t]{0.9\textwidth}
        \centering
        \includegraphics[width=\linewidth]{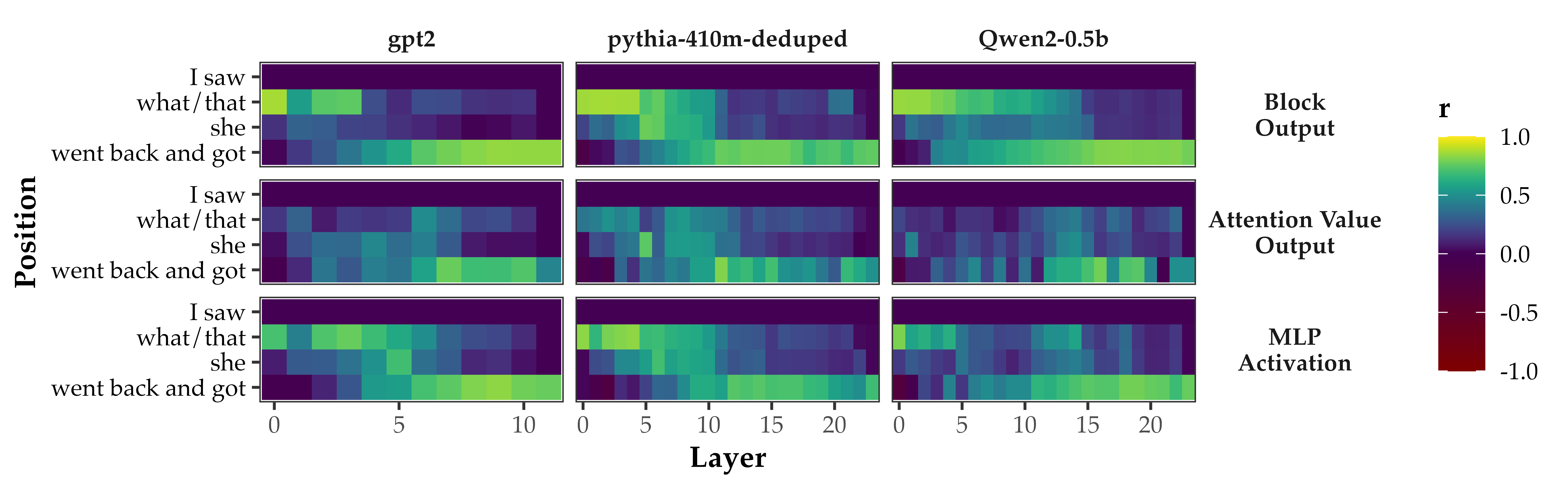}
        \caption{Pearson $r$ correlation  between \textbf{\textsc{odds}} for `classic' filler-gap interventions performed on conjunct stimuli and mean licensing-score. We see high correlation, suggesting greater generalization of mechanisms to `extractable' conjuncts.}
        \label{fig:generalization-bottom}
    \end{subfigure}

    \caption{We use DAS to find the mechanisms to process embedded \textit{wh}-questions before measuring the generalization of said mechanisms to our conjunct stimuli.}
    \label{fig:generalization-figure}
    \vspace{-1.5em}
\end{figure*}

\section{Exp. 2: LMs Process Conjuncts with `Classic' Filler-Gap Mechanisms}\label{sec:experiment-two}

To find the mechanisms LMs use to process extraction (or lack thereof) in these conjuncts, we use the methodology of \citet{boguraev-etal-2025-causal} to probe generalization between constructions. That is, we train DAS on non-conjunct embedded \textit{wh}-question sentences (e.g., Example \ref{ex:fg}) to find interventions causally responsible for mediating extraction. We then measure the causal efficacy of these interventions on our conjunct stimuli (Example \ref{ex:conjunct-island}). The more efficacious these interventions, the stronger the evidence for shared processing mechanisms.

{\exampleindent=2em
\noindent
\begin{minipage}[t]{0.4\linewidth}
\begin{examples}
  \item
    {
    \begin{examples}
      \item I saw \textbf{what} she got $\Rightarrow$ \textbf{.} \label{ex:2a}
      \item I saw \textbf{that} she got $\Rightarrow$ \textbf{the} \label{ex:2b}
    \end{examples}
    }\label{ex:fg}
\end{examples}
\end{minipage}%
\hfill
\begin{minipage}[t]{0.6\linewidth}
\begin{examples}
  \item
    {
    \begin{examples}
      \item I saw \textbf{what} she went back and got $\Rightarrow$ \textbf{.} \label{ex:3a}
      \item I saw \textbf{that} she went back and got $\Rightarrow$ \textbf{the} \label{ex:3b}
    \end{examples}
    }\label{ex:conjunct-island}
\end{examples}
\end{minipage}
}

\paragraph{Methods} We sample training sets of embedded \textit{wh}-questions without coordinated verb-phrases and train DAS as described in \cref{sec:interp-setup}.
We first validate these alignments on a held-out test set of 100 minimal pairs sampled from the same distribution as the training set before evaluating them on 100 minimal pairs for each of our 33 conjuncts to see how much the mechanisms generalize. 
We then compute the mean correlation across seeds between the resulting \textbf{\textsc{Odds}} for each conjunt and the corresponding \textit{wh}-licensing.

\paragraph{Results} Figure \ref{fig:generalization-middle} illustrates the \textbf{\textsc{Odds}} of the learned `classic' filler-gap interventions when evaluated on a held-out test set. Across all three intervention sites, we see strong causal efficacy -- suggesting our interventions robustly capture the mechanisms of `classic' filler-gap extraction. We further see that MLPs are involved more ubiquitously in processing, with the attention modules showing strong causal effect at sites preceding MLP efficacy. This suggests that the attention modules are generally responsible for moving information \citep[consistent with][]{elhage2021mathematical, meng2022locating}, with the MLPs responsible for linguistic processing of that moved information.
Figure \ref{fig:generalization-bottom} further shows high correlation across all positions of the conjuncts between the \textit{wh}-interaction score and \textbf{\textsc{Odds}}.

\paragraph{Discussion} Taken together, our results demonstrate that the more saliently extraction is represented by an LM, the more `classic' filler-gap mechanisms generalize to it. That is, when an LM predicts extraction from a VP conjunct, it makes that prediction through the same pathways it uses to process extraction from `classic' embedded \textit{wh}-questions. However, in cases in which the LM does not predict extraction from a coordinated verb-phrase, this `classic' mechanism gets blocked, only permitting extraction to a gradient extent.

\begin{figure*}[t]
    \centering

    \begin{subfigure}{0.9\textwidth}
        \centering
        \includegraphics[width=\linewidth]{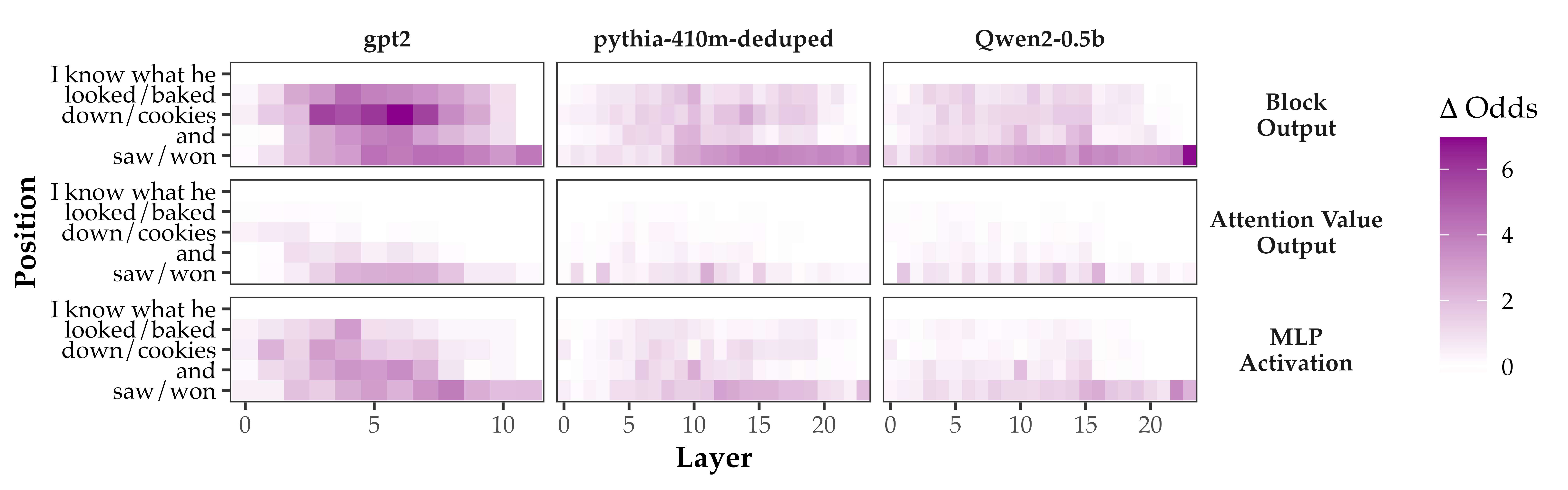}
        \caption{\textbf{\textsc{Odds}} for learned interventions at each position of the conjuncts and layer of every LM when evaluated on a held-out test set. We see strong causal efficacy suggesting DAS has robustly discovered the associated blocking mechanism.}
        \label{fig:blocking-odds}
    \end{subfigure}
    \vspace{-.55em}
    \begin{subfigure}{0.9\textwidth}
        \centering
        \includegraphics[width=\linewidth]{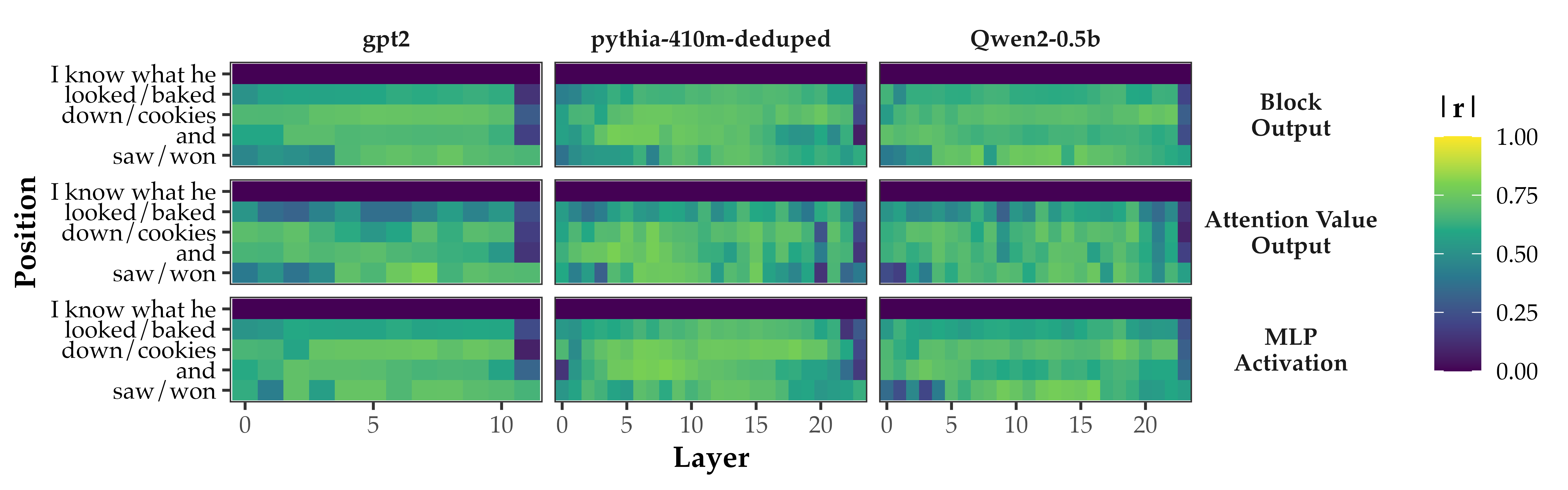}
        \caption{Absolute correlation between each conjunct's average position along the learned subspaces and \textit{wh}-licensing. We see strong generalization of these mechanisms to all conjuncts along the gradients of behavior, despite being trained on a subset of conjuncts.}
        \label{fig:blocking-generalization}
    \end{subfigure}

    \caption{We use DAS to find the causal drawbridges responsible for stranding and unstranding coordination islands.} 
    \label{fig:blocking}
    \vspace{-1em}
\end{figure*}

\section{Exp. 3: LM Judgments are Modulated by Causal Drawbridges}\label{sec:blocking-das}

To find the mechanisms responsible for LMs' gradient behavior, we again use DAS. We learn interventions between conjuncts with strongly `acceptable' and `unacceptable' extractions. In doing so, we aim to find the causal drawbridges which strand and unstrand these syntactic islands. Recognizing that faithful mechanisms must explain the LMs' full behavior, we evaluate these learned interventions on held-out, gradiently acceptable conjuncts. The degree to which these mechanisms reflect gradient behavior serves as a measure for faithfulness.

\paragraph{Methods}
We utilize a set of 16 conjuncts: 8 deemed highly acceptable by both humans and LMs and 8 similarly deemed highly unacceptable. A full list of conjuncts, and acceptability ratings, are in \Cref{app:islands-stimuli}. For a given experiment, we randomly partition these into 6 conjuncts to train on and 2 for held-out evaluation, yielding different training and validation sets for each seed. We then sample 400 minimal pairs of our train conjuncts, differing in the coordinated VP from which extraction is occurring, and train DAS at each position. Example \ref{ex:blocking} shows exemplar minimal pairs. To evaluate, we sample 100 minimal pairs from the held-out conjuncts on which to validate the efficacy of our interventions. 

\begin{examples}
  \item
    {
    \begin{examples}
      \setlength{\exampleindent}{1em}
      \setlength{\labelsep}{0.2em}
      \item I saw what she \textbf{looked down and saw} $\Rightarrow$ \textbf{.} \label{ex:7a}
      \item *I saw what she \textbf{baked cookies and won} $\Rightarrow$ \textbf{the} \label{ex:7b}
    \end{examples}
    }\label{ex:blocking}
\end{examples}

\noindent 
DAS minimal pairs are typically sampled with each class of minimal pair as base and source during training. However, due to one of our minimal pairs being necessarily ungrammatical (i.e., Example 7\ref{ex:7b} has no grammatical continuation), we fix our `unextractable' stimuli as our base and `extractable' variants as our source. In doing so, DAS learns subspaces which cause models to predict `unacceptable' stimuli to be `acceptable'.

To measure whether these subspaces faithfully represent an LM's gradient behavior, we sample evaluation sets of 100 stimuli for all conjuncts, including those not seen during training, and measure their position on these subspaces. If our interventions faithfully represent the LM's `blocking' mechanisms, not only would `acceptable' and `unacceptable' stimuli fall on the subspace's extremes, but the gradient items would fall correspondingly in the middle. To this end, we measure the absolute correlation between a conjunct's subspace position and its \textit{wh}-licensing interaction. A high correlation constitutes evidence that our interventions faithfully represent the mechanisms responsible for an LM's gradience. 

Our hypothesis is that the mechanism for blocking will be primarily localized to the second verb, since that is the point at which the extractability becomes clear.

\paragraph{Results}
Our results suggest these interventions capture LMs' gradient behavior. That is, \Cref{fig:blocking-odds} shows the interventions capture causally efficacious information flow through the LMs and \Cref{fig:blocking-generalization} suggests these interventions generalize across conjuncts.
Interestingly we observe causal effects (to varying extents) starting at the \textit{first} verb position, suggesting relevant information is available as soon as the first verb. 
Post-hoc, we believe this is because the first verb in extractable conjuncts tend to be qualitatively different (e.g., less transitive, semantically lighter) than in unextractable conjuncts. To provide empirical support for these differences, we train a logistic regression on GLoVe embeddings \citep{pennington-etal-2014-glove} of the first verb and the first VP (averaged) to predict extractability. Using leave-one-out validation across the 16 train conjuncts, we find accuracy of 75\% and 100\% respectively, suggesting the verbs (and complements) have meaningfully different character.

We also see qualitatively different mechanisms across LMs. That is, while \texttt{pythia-410m-deduped} and \texttt{Qwen2-0.5b} show weak signal until the final position, \texttt{gpt2} shows strong efficacy in all positions from the first verb's complementizer all the way through the last verb. This suggests that \texttt{gpt2} is more heavily reliant on cues from the first VP
However, that we see these mechanisms to be strongly explanatory of the gradient behavior (high absolute \textit{r} across positions) suggests that the LMs may employ incremental processing from early on in the conjunct, with the different relative efficacies suggesting that this is employed to different degrees. This mirrors psycholinguistic accounts of incremental processing, where verbs constrain downstream reference resolution \citep{altmann1999incremental} and could be tested in humans using eye-tracking experiments testing how the lexical semantics of the first VP constrain predictions on extraction.

\paragraph{Discussion} These results show evidence of causal drawbridges: mechanisms that modulate the gradient acceptability of filler-gap extractions. In the next section, we show these causal drawbridges can help identify related constructions, providing linguistic insights.\footnote{For similar results and discussion correlating subspace location with human acceptability see \cref{app:human-corr}.}

\section{Exp. 4: Learned Subspaces Represent Linguistic Features}

 So far, we have found the abstract mechanisms (i.e., subspaces in an LM's latent space) responsible for gradient extraction across conjuncts. However, this has revealed few linguistic insights. Our last experiment explores the linguistic features encoded in such subspaces.

\paragraph{Methods} 
We focus on qualitative analyses of the subspaces learned in Section \ref{sec:blocking-das}, which constitute the 'blocking mechanism'. To do so, we chunk the Gutenberg Corpus \citep[an open-source English corpus;][]{gutenberg} into segments matching the length of our stimuli (8 tokens). We then randomly select 100,000 chunks, collect the activations of an LM processing those chunks, and project those activations onto the relevant subspaces. By analyzing the chunks' positions within these subspaces, we aim to gain insight into what properties they encode and why they are causally relevant for LM behavior.
 
Our method is akin to work associating features with SAEs and Transcoders through qualitative analysis of a corpus' top activating examples \cite{cunningham2023sparse, bricken2023monosemanticity, dunefsky2024transcoders}. The difference is that other methods find features which take on a value in a range from ``activated'' to ``not activated'', which have been trained in an unsupervised manner to represent singular features. In contrast, our method identifies gradable features defining causally-relevant subspaces found through supervised training.

We focus on \texttt{gpt2}, specifically on a highly efficacious intervention in the Transformer Block Output at the \textsc{and} position. While the second verb is not yet available at this site, the results in \cref{sec:blocking-das} show that information relevant for extraction is already available. This experiment is an attempt to understand which linguistic features the LM is tracking.
We do not claim to have reverse-engineered the full blocking mechanism, leaving this to future work. To that end, we release all chunks and corresponding values at every site of every model.\footnote{Chunks and accompanying visualizer can be found at \href{http://island-chunks.s3-website-us-east-1.amazonaws.com/chunk_viz.html}{this link}. } We stress that this analysis is qualitative and noisy: the subspace was learned to discriminate extractable from unextractable conjuncts, but we cannot directly observe what linguistic features drive that discrimination, nor rule out surface correlates such as transitivity.

\begin{table*}[t]
\footnotesize
\centering
\setlength{\tabcolsep}{2pt}
\renewcommand{\arraystretch}{1.1}
\resizebox{\textwidth}{!}{%
\begin{tabular}{@{}lcllll@{}}
\toprule
\textbf{Position} &
\textbf{Layer} &
\textbf{$\Delta$\textsc{Odds}} &
\textbf{$|r|$} & 
\textbf{`Extractable' Examples} &
\textbf{`Unextractable' Examples}\\
\midrule
\textsc{and}&6 & 5.03 & 0.67 & ...his life he had crept up \textcolor{ForestGreen}{\textbf{and}} stolen...& ...you men, we offer you meat \textcolor{BrickRed}{\textbf{and}} bread... \\
   && & & ...valleys, so that men look up \textcolor{ForestGreen}{\textbf{and}} see... & ...the groom’s books \textcolor{BrickRed}{\textbf{and}} papers...\\
   &&  &  & ...I think we're in \textcolor{ForestGreen}{\textbf{for}} a... & ...offers on the best terms for cash \textcolor{BrickRed}{\textbf{or}} approved... \\
   && & & ...it in motion, she started up \textcolor{ForestGreen}{\textbf{and}} said... & ...chief took a spoonful of meat \textcolor{BrickRed}{\textbf{and}} liquor...\\
   && & & ...ieved in and preached what he had \textcolor{ForestGreen}{\textbf{just}} heard... & ...as pigs, fowl, dogs\textcolor{BrickRed}{\textbf{,}} cattle... \\
   &&  &  & ...ourd Wood, we sat down \textcolor{ForestGreen}{\textbf{to}} make... &  ...a. I hab no wife \textcolor{BrickRed}{\textbf{nor}} child...\\
   && & & ...dozen willing hands reached down \textcolor{ForestGreen}{\textbf{to}} catch... &...A broken chair, a stool \textcolor{BrickRed}{\textbf{or}} two... \\
\bottomrule
\end{tabular}
}
\caption{Exemplar chunks activated near `extractable' and `unextractable' conjuncts in the Transformer Block Output of \texttt{gpt2}. 
Full lists of top 75 such chunks at this site are in \cref{app:chunks}.}
\label{tab:chunks}
\vspace{-1.5em}
\end{table*}

\paragraph{Results} 
We see many chunks with \textit{and}, likely given the lexical nature of our conjuncts. However, the \textit{and} tokens near the `extractable' and `unextractable' items posses different senses. In particular, the \textit{and} near extractable items seems akin to an intra-event connective, conjoining verb-phrases (such as \textit{crept up and stolen}, \textit{look up and see}...). Conversely, in the `unextractable' examples, the \textit{and} functions as a logical conjunction, conjoining nominal entities or conceptual categories (e.g., \textit{meat and bread}, \textit{books and papers}...). 

However, it is particularly instructive to examine activated ``non-and'' tokens. We see a trend in which tokens fall where on the subspaces, with \textit{to}, acting as a purpose marker, \textit{for}, expressing a causal consequent, and \textit{just}, marking immediate precedence, activated in the `extractable' case. And in contrast, in the `unextractable' case, logical disjunctions like \textit{or}, \textit{nor}, and conjunctival punctuation are activated. 

\paragraph{Discussion} These results highlight a novel hypothesis: in unextractable cases the conjunction behaves akin to a logical conjunction, but in extractable cases it takes on the character of function words encoding relational dependencies.
This hypothesis is well situated in the literature. First, LMs are known to be highly sensitive to polysemous usages  \citep{chronis-erk-2020-bishop}, grounding the plausibility of this distinction.
Secondly, this distinction tracks with prior work in theoretical linguistics aimed at explaining CSC violations. 
Ross himself noticed the asynchrony between extractable and unextractable conjuncts and proposed that those permitting extraction, such as \textit{went to the store and bought}, were derived from the underlying structure \textit{went to the store to buy} \citep{ross1967constraints}, and are not in fact conjunctions at all, merely resembling them on the surface. While this analysis has been convincingly argued against \citep{schmerlinc1975asymmetric}, it captures similar intuitions to ours, namely that the conjunction in these cases functions not as purely logical conjunction, but as a purposive or sequential marker. Ross, however, commits strongly to this distinction being a categorical, syntactic one, whereas our analysis suggests this is a gradable property of the conjunction's lexical semantics. Separately, \citet{kehler2002coherence} classifies VP coordination into Parallel relations, where the two coordinated VPs are unrelated, Occasion relations, in which the two VPs are in natural sequence and Cause-Effect relations in which the first VP causes the second VP. He posits that extraction is not permitted from a Parallel relation but is permitted from Occasion and Cause-Effect relations. And accordingly, we see the `extractable' chunks are these such permissible relations, whereas the `unextractable' chunks are not.

We encourage future psycholinguistic work testing this hypothesis, e.g, syntactic priming paradigms measuring whether people process phrases like \textit{went out and got} more akin to activated phrases like \textit{riding up to him} than classical conjuncts like \textit{drank whiskey and ate}.
We are optimistic our results point to further work on the lexical semantics of conjunctions.

\section{Conclusion}

There have been recent claims that LMs aid in developing insights into linguistic structure and representations \citep[e.g.,][]{futrell2025linguistics}. 
Here, we hope to have delivered such insights. We showed that a variety of LMs show human-like sensitivity to the gradient acceptability of supposed syntactic islands, and that these constructions are handled by a more general filler-gap mechanism. We then show that we can use causal interventions to understand the drawbridge mechanism that controls the gradience of acceptability. We end by providing evidence that these interventions can help generate novel linguistic insights. 

\section{Limitations}

We claim that we generate linguistic insights,  to see if those insights generalize to humans,  human experiments would be helpful. We suggest possible directions for human experiments, but such experiments were outside of the scope of this project.

We focus on English constructions here but note that some of the importance of syntactic islands in linguistic history depends on their generalizability across languages -- particularly in the case of the Coordinate Structure Constraint which is claimed to be remarkably robust multi-lingually. Thus, extending this work to non-English languages would enhance its generalizability, particularly given differential performance of models on filler-gaps in non-English languages \citep{kobzeva2023neural}. 

We recognize that the models we tested see more data than human learners see, and should thus moderate any claims about Poverty of Stimulus arguments. At the same time, we also note that the models we test are smaller and weaker than frontier models and so these results do not necessarily reflect either the performance or internal analyses of state-of-the-art LMs, which could differ both in behavioral performance and mechanism.

\section{LLM Usage Statement}
We used Claude to provide feedback on the clarity of our writing during the editing process. We further used Claude to aid in the development of the code used to perform Experiment 4, and to do post-hoc optimization of the code written for Experiments 2 and 3 which allowed for faster experimentation. 

\section{Acknowledgments}

We thank Abigail Fergus and Adele Goldberg for their LSA presentation that introduced us to their data set and for making their data available. We also thank Andy Kehler for his insights and discussion on an initial version of this work.
We acknowledge funding from NSF CAREER grant 2339729 to Kyle Mahowald.

\bibliography{anthology1,anthology2,anthology3,custom,everything,colm2026_conference}

\appendix

\section{Stimuli Conversion}\label{app:conversion}

The stimuli we source from \citet{fergus2025islands} are in the form of matrix wh-questions. This makes minimal pair templates difficult to form. As such, we convert them into embedded wh-questions. Below, we detail this conversion process. 

\begin{examples}
    \item What did she bake cookies and win?
    \label{ex:example_m}
\end{examples}

Given the stimuli in (\ref{ex:example_m}), we convert it into the corresponding embedded question, as in (\ref{ex:example_e}).

\begin{examples}
    \item I know what she baked cookies and won $\rightarrow$ .
    \label{ex:example_e}
\end{examples}

\noindent To facilitate this, we make the stimuli past-tense as well. The example in (\ref{ex:example_e}) then serves as one of our minimal pairs (in particular the one with a gap), with the corresponding minimal pair shown in (\ref{ex:example_e2}).

\begin{examples}
    \item I know that she baked cookies and won $\rightarrow$ the
    \label{ex:example_e2}
\end{examples}

This is just one such minimal pair sample from our templates, with the prefix (in this pair `\textit{I know}') and subject (`\textit{she}') sampled from a large set of potential items to allow for diverse minimal pairs.

We note that such changes have the potential to induce different argument structures, affecting the licensing. However, we believe these effects to be minimal, and expect the acceptability judgments collected by \citet{fergus2025islands} to serve as reliable judgments for the embedded \textit{wh}-questions.

\section{Training and Evaluation Details}\label{app:train-eval-detail}

We access all LMs used in this study through the \texttt{transformers} python package \citep{wolf2020huggingfacestransformersstateoftheartnatural}. In order to calculate LM surprisals for our stimuli, we utilize the \texttt{minicons} library \citep{misra2022minicons}. To train DAS and leverage the resulting interventions, we use the \texttt{pyvene} library \citep{wu-etal-2024-pyvene}.

When training DAS, we train for one epoch on 400 unique minimal pairs, utilizing a batch size of 4, and gradients are accumulated over 4 batches per step. We use the Adam optimizer with a learning rate of $5e-3$ and apply 100 warm-up steps. In the case of token length mismatch of aligned stimuli, we train interventions on the last token of a given span. 

All experiments were run on 1 NVIDIA A40 GPU. Behavioral experiments take approximately 8 hours for all models. 

When training DAS on \texttt{gpt2} each intervention takes approximately 5 seconds per intervention, totaling approximately 20 minutes and 25 minutes per seed for Experiment Two and Three respectively. Training DAS on \texttt{pythia-410m-deduped} takes approximately 12 seconds per intervention, totaling approximately 55 minutes and 75 minutes per seed for Experiment Two and Three respectively. Finally, for \texttt{Qwen2-0.5b} each intervention takes approximately 17 seconds, totaling approximately 100 minutes per seed. 

Evaluation for Experiment Two takes approximately 30 seconds per intervention for \texttt{gpt2}, totaling approximately 45 minutes per seed, 40 seconds per intervention for \texttt{pythia-410m-deduped} totaling approximately 150 minutes per seed, and 60 seconds per intervention for \texttt{Qwen2-0.5b} totaling approximately 210 minutes per seed.

Evaluation for Experiment Three takes approximately 20 seconds per intervention for \texttt{gpt2}, totaling approximately 45 minutes per seed, 20 seconds per intervention for \texttt{pythia-410m-deduped} totaling approximately 90 minutes per seed, and 40 seconds per intervention for \texttt{Qwen2-0.5b} totaling approximately 180 minutes per seed.

Evaluating the chunks in Experiment 4 takes approximately 20 seconds per intervention, totaling approximately 45 minutes per seed for \texttt{gpt2} and 90 minutes per seed for \texttt{pythia-410m-deduped} and \texttt{Qwen2-0.5b}.

\section{Human and LM Behavioral Correlation}
\begin{figure*}[t]
    \centering
    \includegraphics[width=.9\linewidth]{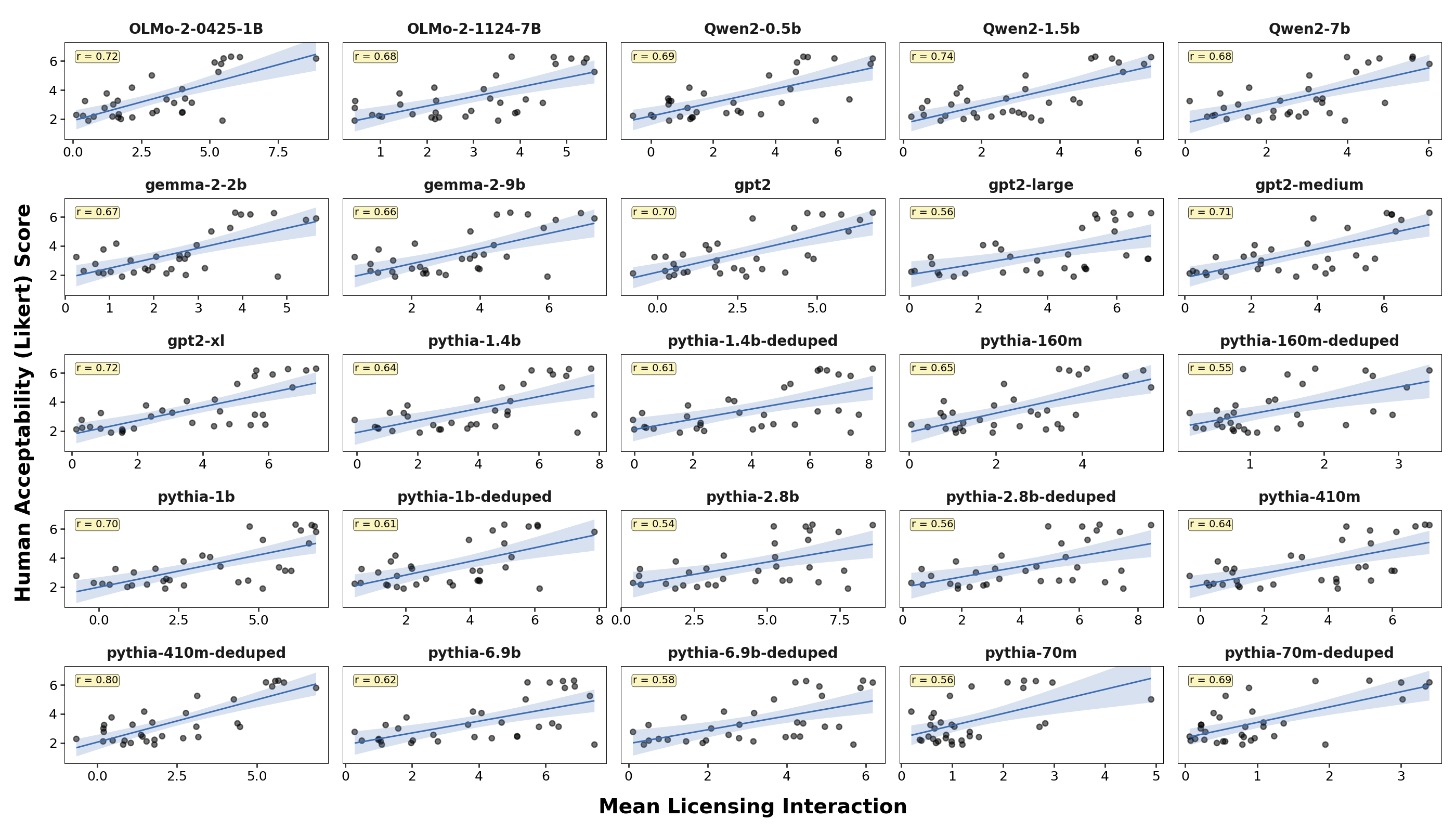}
    \caption{Correlation between LM mean licensing interaction and human acceptability judgments across embedded \textit{wh}-questions with verb-phrase conjuncts. Across a broad class of LM families and sizes, we largely find agreement (high pearson $r$ correlation coefficient) between the two measures. }
    \label{fig:lm-human}
\end{figure*}

 In \Cref{fig:lm-human} we provide a full set scatter plots for each LM displaying the given LM mean licensing interaction plotted as a function of human acceptability rating for each conjunct in our stimuli. We further include the reported Pearson $r$ value for each conjunct. 

\section{DAS Blocking Mechanism Training Data}\label{app:islands-stimuli}
A full list of conjuncts used to train DAS to find the blocking mechanism, along with their respective human acceptability ratings and LM \textit{wh}-licensing scores, can be seen in \Cref{tab:islands}.

\begin{table*}[t]
\footnotesize
\centering
\setlength{\tabcolsep}{2pt}
\renewcommand{\arraystretch}{1.1}
\resizebox{\textwidth}{!}{%
\begin{tabular}{@{}llll|cccccc@{}}
\toprule
\textbf{\textsc{Verb$_1$}} &
\textbf{\textsc{Complement}} &
\textbf{\textsc{And}} &
\textbf{\textsc{Verb$_2$}} &
\textbf{\begin{tabular}{@{}c@{}}Human\\Acceptability\end{tabular}} &
\textbf{\begin{tabular}{@{}c@{}}\texttt{gpt2} Mean \\ \textit{Wh}-Licensing\end{tabular}} &
\textbf{\begin{tabular}{@{}c@{}}\texttt{pythia-410m-ded.} \\ Mean \textit{Wh}-Licensing\end{tabular}} &
\textbf{\begin{tabular}{@{}c@{}}\texttt{Qwen2-0.5b} Mean \\ \textit{Wh}-Licensing\end{tabular}} &
\textbf{Class}\\
\midrule
  \midrule        
  looked & down & and & saw & 6.29 & 6.75 & 5.68 & 4.90 & \textcolor{ForestGreen}{Acceptable}\\         
  went & home & and & got & 6.26 & 4.70 & 5.58 & 5.04 & \textcolor{ForestGreen}{Acceptable}\\
  woke & up & and & smelled & 6.16 & 5.76 & 5.85 & 7.13 & \textcolor{ForestGreen}{Acceptable}\\
  came & back & and & talked about & 6.17 & 5.17 & 5.29 & 5.89 & \textcolor{ForestGreen}{Acceptable}\\
  ran & out & and & got & 5.89 & 2.99 & 5.48 & 4.69 & \textcolor{ForestGreen}{Acceptable}\\
  woke & up & and & thought of & 5.79 & 6.35 & 6.86 & 7.06 & \textcolor{ForestGreen}{Acceptable}\\
  drove & to the market & and & bought & 5.24 & 4.29 & 3.13 & 4.66 & \textcolor{ForestGreen}{Acceptable}\\
  sat & back & and & enjoyed & 5.00 & 5.99 & 4.28 & 3.80 & \textcolor{ForestGreen}{Acceptable}\\
  \midrule
  picked & strawberries & and & wrote & 1.89 & 0.37 & 0.82 & 0.59 & \textcolor{BrickRed}{Unacceptable}\\
  loved & painting & and & enrolled in & 2.00 & 0.53 & 1.05 & 1.27 & \textcolor{BrickRed}{Unacceptable}\\
  baked & cookies & and & won & 2.16 & 0.83 & 0.86 & 0.08 & \textcolor{BrickRed}{Unacceptable}\\
  ate & candy & and & developed & 2.11 & -0.76 & 0.19 & 1.34 & \textcolor{BrickRed}{Unacceptable}\\
  read & a magazine & and & enrolled in & 2.11 & 1.97 & 1.55 & 1.31 & \textcolor{BrickRed}{Unacceptable}\\
  played & soccer & and & planted & 2.22 & 0.95 & 1.80 & -0.57 & \textcolor{BrickRed}{Unacceptable}\\
  called & his boyfriend & and & painted & 2.28 & 0.25 & -0.65 & 0.01 & \textcolor{BrickRed}{Unacceptable}\\
  cooked & dinner & and & read & 2.76 & 0.51 & 0.21 & 1.17 & \textcolor{BrickRed}{Unacceptable}\\
\bottomrule
\end{tabular}
}
\caption{\textbf{Left Block:} The coordinated clause of the coordination-islands used to train DAS on the blocking mechanism. The top items are `acceptable' coordination-islands (strong preference for extraction), and the bottom four `unacceptable' (strong lack of preference for extraction).
\textbf{Right Block}: Human Acceptability scores along a Likert scale and mean licensing-scores for each LM we mechanistically evaluate. We use conjuncts which \textbf{a)} broadly demonstrate calibration between human and LM judgments and \textbf{b)} have length-matched tokens.}
\label{tab:islands}
\end{table*}

\section{Subspaces in LM Hidden Spaces Correlate with Human Judgments}\label{app:human-corr}

We replicate the analysis of the blocking mechanism as in Section \ref{sec:blocking-das}, but instead correlate conjunct position on the subspace with human acceptability judgments rather than LM \textit{wh}-licensing score. 

\begin{figure*}[ht]
        \centering
        \includegraphics[width=\linewidth]{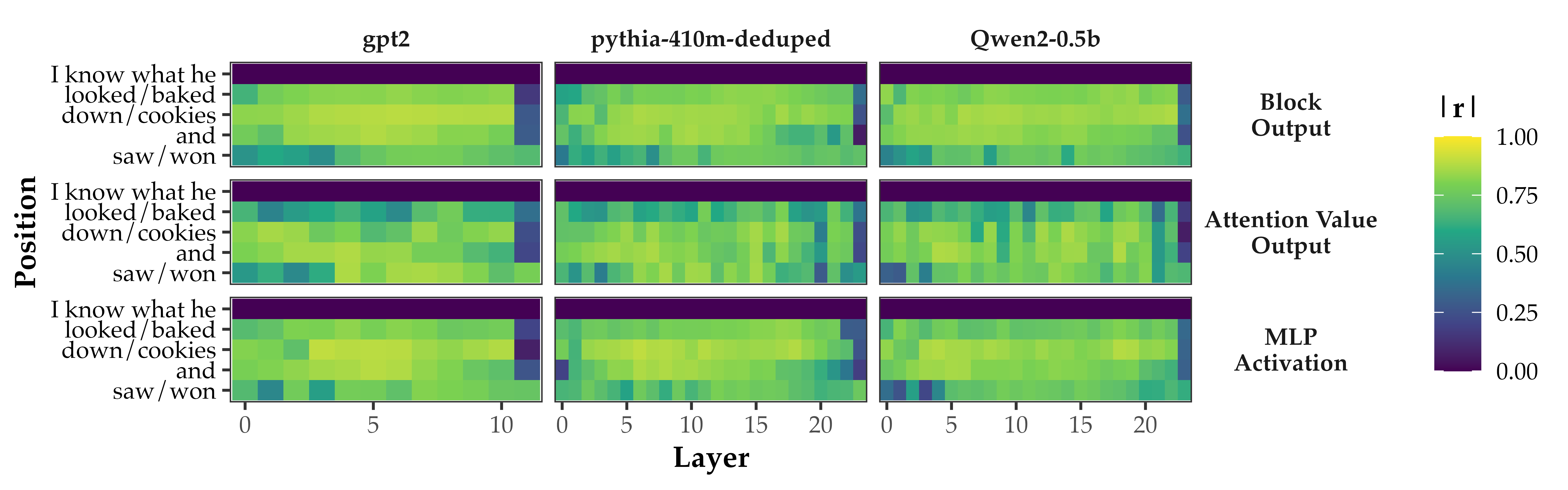}
        \caption{Absolute correlation between each conjunct's average position along the learned subspaces (on which an intervention is performed) and human judgments. We see strong correlation between position on subspace and human judgments.}
        \label{fig:blocking-generalization-human}
\end{figure*}

Our results can be seen in Figure \ref{fig:blocking-generalization-human}. Interestingly, we find extremely strong correlations across the majority of positions -- notably higher than the correlation with LM \textit{wh}-licensing score shown in Figure \ref{fig:generalization-bottom}. While we do not investigate these results further, we find them extremely exciting, and in line with prior work which has shown that LLM internals align strongly with human behavior \citep{kuribayashi2025large}. We see these as preliminary results suggesting that there are subspaces, discoverable through methods like DAS, which strongly correlate LM internals with human behavior. Future work can build upon these results, discovering further behavior correlating with LMs internal subspaces and localizing where specifically in the models this correlation occurs.

\section{Chunk Extraction Protocol}
In Experiment 3, we report the absolute Pearson-r due to the fact that the position on a subspace can correlate either positively or negatively with \textit{wh}-licensing. This is because the interventions are merely optimized such that intervening on the subspace changes the model's output accordingly, and this is agnostic to which side of a subspace each class of a minimal pair gets seperated to. Due to this fact, it is non-trivial to take the top-activating chunks averaged over the five seeds. Our solution to this problem is once we have calculated the correlation with model \textit{wh}-licensing, we multiply all activations by the sign of that correlation, before normalizing by z-score. We then average across the resulting normalized activations.

\section{Top-Activating Chunks}\label{app:chunks}
We provide the top 75 most activated chunks at the given position we study, averaged across the five trained sets of interventions for \texttt{gpt2} at the 
\textsc{And} positions in Table \ref{tab:chunk-gpt2}
. Note that we filter out the chunks for which the token activated at a given position was whitespace for heightened clarity of results. 

\begin{table*}[t]
\centering
\scriptsize
\setlength{\tabcolsep}{4pt}
\begin{tabular}{r|p{0.4\textwidth}|p{0.5\textwidth}}
\toprule
 & \textbf{Top `Extractable' Examples} & \textbf{Top `Unextractable' Examples} \\
\midrule
1 & \textbackslash nhis life he had crept up\textcolor{ForestGreen}{\textbf{ and}} stolen &  you men, we offer you meat\textcolor{red}{\textbf{ and}} bread \\
2 & orsyth and Ronald were walking back\textcolor{ForestGreen}{\textbf{ and}} forth &  partly to discover the places where gold\textcolor{red}{\textbf{ and}}\textbackslash r \\
3 &  her room she saw him pacing back\textcolor{ForestGreen}{\textbf{ and}} forth &  the eating of animal food, wine\textcolor{red}{\textbf{ and}} marriage \\
4 &  quilt. Her fingers went in\textcolor{ForestGreen}{\textbf{ and}} out & \textbackslash nyoung men how he threw spears\textcolor{red}{\textbf{ and}} har \\
5 & \textbackslash nUneasy wanders to\textcolor{ForestGreen}{\textbf{ and}} fro &   the groom's books\textcolor{red}{\textbf{ and}} papers \\
6 & \textbackslash r\textbackslash n\textbackslash r\textbackslash nSo far,\textcolor{ForestGreen}{\textbf{ so}} clear &  the most nutritive of all meat\textcolor{red}{\textbf{ and}}\textbackslash r \\
7 & ! Churn!\textbackslash r\textbackslash nUp\textcolor{ForestGreen}{\textbf{ and}} down &  a lot of old muskets\textcolor{red}{\textbf{ and}} ten \\
8 &  think, one has to be up\textcolor{ForestGreen}{\textbf{ and}}\textbackslash r & ast assortment of duodec bombs\textcolor{red}{\textbf{ and}} other \\
9 &  valleys, so that men look up\textcolor{ForestGreen}{\textbf{ and}} see &  representing tomatoes, radishes, apples\textcolor{red}{\textbf{,}} p \\
10 & cher and his lanky friend entered\textcolor{ForestGreen}{\textbf{ and}} draped & and Mère Marie found some pots\textcolor{red}{\textbf{ and}}\textbackslash r \\
11 & . Perhaps some one will come down\textcolor{ForestGreen}{\textbf{ and}} show & \textbackslash n    sort of grammar\textcolor{red}{\textbf{ and}} sense \\
12 & \textbackslash n\textbackslash r\textbackslash nbeer the world has\textcolor{ForestGreen}{\textbf{ ever}} known & \textbackslash r\textbackslash n\textbackslash r\textbackslash ndesirable dogs\textcolor{red}{\textbf{ and}} his \\
13 &  it in motion, she started up\textcolor{ForestGreen}{\textbf{ and}} said &  baking of the squaw corn cakes\textcolor{red}{\textbf{ and}} the \\
14 & When the village was passed I stood\textcolor{ForestGreen}{\textbf{ and}} looked &  an egg, a bit of bacon\textcolor{red}{\textbf{,}} ordered \\
15 & \textbackslash r\textbackslash n\textbackslash r\textbackslash nreached up\textcolor{ForestGreen}{\textbf{ and}} tried & whatever about the different kinds of trees\textcolor{red}{\textbf{ and}} he \\
16 &  she had hoped. He got up\textcolor{ForestGreen}{\textbf{ and}} carried &  not tell you how glad my wife\textcolor{red}{\textbf{ and}} sons \\
17 & !"\textbackslash r\textbackslash nThe Lieutenant staggered out\textcolor{ForestGreen}{\textbf{ and}} almost &       of heroes\textcolor{red}{\textbf{ and}} law \\
18 & At last, at last we turned\textcolor{ForestGreen}{\textbf{ and}} stood &  me, he clutched both rifle\textcolor{red}{\textbf{ and}} belt \\
19 & \textbackslash nMother Tiger, then he went\textcolor{ForestGreen}{\textbf{ and}} hid &  of them left his Bible, cane\textcolor{red}{\textbf{ and}} hat \\
20 &  to wander\textbackslash r\textbackslash n\textbackslash r\textbackslash nup\textcolor{ForestGreen}{\textbf{ and}} down &    which I put some wool\textcolor{red}{\textbf{ and}} moss \\
21 &  O and the T up-and\textcolor{ForestGreen}{\textbf{-}}down & \textbackslash nmountain. They used spells\textcolor{red}{\textbf{ and}} magic \\
22 & ishing their hot stew! Far down\textcolor{ForestGreen}{\textbf{ and}} nearer & arkan, built of whalebone\textcolor{red}{\textbf{ and}} rubber \\
23 &  village was passed I stood and looked\textcolor{ForestGreen}{\textbf{ back}}. & \textbackslash r\textbackslash n\textbackslash r\textbackslash n\textbackslash r\textbackslash nFood\textcolor{red}{\textbf{ and}} the \\
24 &  if he wakes at midnight he rises\textcolor{ForestGreen}{\textbf{ and}} begins &  vapor of iodine, bromine\textcolor{red}{\textbf{,}} or \\
25 &  Only."\textbackslash r\textbackslash nI went to\textcolor{ForestGreen}{\textbf{ bed}} at & \textbackslash r\textbackslash n\textbackslash r\textbackslash nWisdom literature\textcolor{red}{\textbf{ and}} Phil \\
26 &  again next moment, and I got\textcolor{ForestGreen}{\textbf{ hold}} of &  are constantly spoken of as heretics\textcolor{red}{\textbf{ and}} false \\
27 &  I think we're in\textcolor{ForestGreen}{\textbf{ for}} a & ers on the best terms for cash\textcolor{red}{\textbf{ or}} approved \\
28 & \textbackslash r\textbackslash n\textbackslash r\textbackslash ncame forth\textcolor{ForestGreen}{\textbf{ and}} stood & !' And she brought wood\textcolor{red}{\textbf{ and}}\textbackslash r \\
29 &  thought\textbackslash r\textbackslash n\textbackslash r\textbackslash nwould cease\textcolor{ForestGreen}{\textbf{ and}} determine &  soja beans, beggarweed\textcolor{red}{\textbf{,}}\textbackslash r \\
30 & \textbackslash r\textbackslash nto youth, leading through\textcolor{ForestGreen}{\textbf{ and}} beyond & \textbackslash ntwice as much real education\textcolor{red}{\textbf{ and}} mental \\
31 & \textbackslash n\textbackslash r\textbackslash nthey all sat down\textcolor{ForestGreen}{\textbf{ and}} waited &       my breast\textcolor{red}{\textbf{ and}} yelled \\
32 & \textbackslash r\textbackslash nMr. McCool goes\textcolor{ForestGreen}{\textbf{ and}} gives & \textbackslash r\textbackslash nshould be given nearby oak\textcolor{red}{\textbf{ and}} h \\
33 &   The green posters were distributed far\textcolor{ForestGreen}{\textbf{ and}} wide & handles, broom-sticks,\textcolor{red}{\textbf{ and}} s \\
34 &  and shivering, they stumbled outside\textcolor{ForestGreen}{\textbf{ and}} by &  the roar of its hundred heavy guns\textcolor{red}{\textbf{ and}} mort \\
35 & ieved in and preached what he had\textcolor{ForestGreen}{\textbf{ just}} heard & \textbackslash ngrain, and introducing clover\textcolor{red}{\textbf{ and}} new \\
36 &  corn for you.\textbackslash r\textbackslash nCome\textcolor{ForestGreen}{\textbf{ and}} eat & 17.A great deal of rain\textcolor{red}{\textbf{ and}} very \\
37 & \textbackslash r\textbackslash n\textbackslash r\textbackslash nTO THE IND\textcolor{ForestGreen}{\textbf{IF}}FER &  four-lobed calyx\textcolor{red}{\textbf{ and}} sometimes \\
38 &  please.\textbackslash r\textbackslash n``There\textcolor{ForestGreen}{\textbf{ you}} are & ur Gilson. Take the food\textcolor{red}{\textbf{ and}}\textbackslash r \\
39 & . Antwerp will fall to\textcolor{ForestGreen}{\textbf{-}}morrow & \textbackslash n\textbackslash r\textbackslash nbloated flesh\textcolor{red}{\textbf{ and}} redd \\
40 & \textbackslash r\textbackslash nThis was refreshing doctrine to\textcolor{ForestGreen}{\textbf{ come}} from &  ponderous, helpless knowledge of books\textcolor{red}{\textbf{,}} for \\
41 &  for the thought.\textbackslash r\textbackslash nNow\textcolor{ForestGreen}{\textbf{ and}} then &  spoonful of beef-tea\textcolor{red}{\textbf{,}}\textbackslash r \\
42 & 'm getting paid for." He paused\textcolor{ForestGreen}{\textbf{ and}} removed & \textbackslash n\textbackslash r\textbackslash nthe poor Princes\textcolor{red}{\textbf{ and}} put \\
43 & \textbackslash ndispatch, the Press got\textcolor{ForestGreen}{\textbf{ hold}} of &  chief took a spoonful of meat\textcolor{red}{\textbf{ and}} liquor \\
44 &  list in her hand, which she\textcolor{ForestGreen}{\textbf{ proceeded}} to & \textbackslash r\textbackslash n\textbackslash r\textbackslash nmade admirable sailors\textcolor{red}{\textbf{ and}} excellent \\
45 &  in despair.\textbackslash r\textbackslash nShe turned\textcolor{ForestGreen}{\textbf{ round}}. &  as pigs, fowl, dogs\textcolor{red}{\textbf{,}} cattle \\
46 & essential."\textbackslash r\textbackslash n"You have\textcolor{ForestGreen}{\textbf{,"}} agreed & -dress of dark-blue cloth\textcolor{red}{\textbf{ and}} a \\
47 &  turned his brain that he went home\textcolor{ForestGreen}{\textbf{ and}} determined & subsistence. Turtles, crabs\textcolor{red}{\textbf{,}} p \\
48 & .\textbackslash r\textbackslash n"What are you\textcolor{ForestGreen}{\textbf{ going}} to & .\textbackslash r\textbackslash n\textbackslash r\textbackslash nBirds\textcolor{red}{\textbf{ and}} flowers \\
49 & \textbackslash r\textbackslash nattention. There is\textcolor{ForestGreen}{\textbf{,}} first &  mean an inordinate desire for drink\textcolor{red}{\textbf{,}} a \\
50 &  unselfish!' She turned,\textcolor{ForestGreen}{\textbf{ and}} caught & a. I hab no wife\textcolor{red}{\textbf{ nor}} child \\
51 & nd ed. © 15Sep50\textcolor{ForestGreen}{\textbf{;}} A &   A broken chair, a stool\textcolor{red}{\textbf{ or}} two \\
52 &  place."\textbackslash r\textbackslash nHe looked up\textcolor{ForestGreen}{\textbf{,}} and & , a gold watch-chain,\textcolor{red}{\textbf{ and}} a \\
53 & \textbackslash nwhen his followers came riding up\textcolor{ForestGreen}{\textbf{ to}} him & \textbackslash n\textbackslash r\textbackslash n"Only more hills\textcolor{red}{\textbf{ and}} more \\
54 &  to England--Letters of End\textcolor{ForestGreen}{\textbf{ym}}ion &  deep in grass, with ivy\textcolor{red}{\textbf{ and}} b \\
55 &  own concerns, and leave him und\textcolor{ForestGreen}{\textbf{ist}}urbed & otton and linen rags, hemp\textcolor{red}{\textbf{,}} woods \\
56 & \textbackslash nWhen the men and women came\textcolor{ForestGreen}{\textbf{ home}} that &  Lyons, the country of silks\textcolor{red}{\textbf{ and}}\textbackslash r \\
57 & \textbackslash r\textbackslash n\textbackslash r\textbackslash nI have little\textcolor{ForestGreen}{\textbf{ to}} say & \textbackslash r\textbackslash nby smooth flint stones\textcolor{red}{\textbf{ and}} gains \\
58 &  saw the dark bodies whirled round\textcolor{ForestGreen}{\textbf{ and}} round & \textbackslash n\textbackslash r\textbackslash nsuffered from hunger\textcolor{red}{\textbf{ and}} thirst \\
59 & aham. © 20Mar47\textcolor{ForestGreen}{\textbf{;}} A & \textbackslash n      grain\textcolor{red}{\textbf{ and}} fed \\
60 & .  I had to go to\textcolor{ForestGreen}{\textbf{ bed}} for & \textbackslash nevery day. They were carrots\textcolor{red}{\textbf{,}} of \\
61 &  end of the still figure and turned\textcolor{ForestGreen}{\textbf{ back}} the &  might have been worse, the linen\textcolor{red}{\textbf{ and}} plate \\
62 & ourd Wood, we sat down\textcolor{ForestGreen}{\textbf{ to}} make & \textbackslash r\textbackslash ndrops of bromine\textcolor{red}{\textbf{,}} the \\
63 & ed cadence art could ne'\textcolor{ForestGreen}{\textbf{er}} attain &  greater importance in education, the classics\textcolor{red}{\textbf{ or}}\textbackslash r \\
64 & ---I thought when I came down\textcolor{ForestGreen}{\textbf{ to}}\textbackslash r & ism, anticipating the spirit of Christianity\textcolor{red}{\textbf{ and}} the \\
65 & \textbackslash nAnd the next day he came\textcolor{ForestGreen}{\textbf{ back}}, & \textbackslash nmeteorite-screens\textcolor{red}{\textbf{ and}} wall \\
66 & .\textbackslash r\textbackslash nSimon said unent\textcolor{ForestGreen}{\textbf{hus}}i & geons, apothecaries,\textcolor{red}{\textbf{ and}} physicians \\
67 &  Autry. © 30Aug50\textcolor{ForestGreen}{\textbf{;}} AA & \textbackslash r\textbackslash nwith Indians, Kurds,\textcolor{red}{\textbf{ and}} a \\
68 & \textbackslash nThe Miss Sahiba paused,\textcolor{ForestGreen}{\textbf{ and}} all & out on prejudice or anti-British\textcolor{red}{\textbf{ or}}\textbackslash r \\
69 & . She opened the gate, and\textcolor{ForestGreen}{\textbf{ came}} up &  bromine, iodus,\textcolor{red}{\textbf{ and}} i \\
70 & Wellard. © 26Jan50\textcolor{ForestGreen}{\textbf{;}} A &  of\textbackslash r\textbackslash n\textbackslash r\textbackslash nsoldiers\textcolor{red}{\textbf{ and}} of \\
71 & -sacrifice.  There came\textcolor{ForestGreen}{\textbf{ to}} men & indulge hope for the church\textcolor{red}{\textbf{ and}} the \\
72 &  of the time. I didn't\textcolor{ForestGreen}{\textbf{}} & body there depended a decomposition blaster\textcolor{red}{\textbf{ and}} an \\
73 &  my wife, Fritz had gone down\textcolor{ForestGreen}{\textbf{ to}} his &  of hairline, width of cheeks\textcolor{red}{\textbf{ and}} height \\
74 &  by Pa-khar or Ne\textcolor{ForestGreen}{\textbf{hes}}i & \textbackslash n\textbackslash r\textbackslash nthe hum of insects\textcolor{red}{\textbf{ and}} the \\
75 & \textbackslash r\textbackslash ndozen willing hands reached down\textcolor{ForestGreen}{\textbf{ to}} catch & \textbackslash r\textbackslash nMothers and Wives\textcolor{red}{\textbf{ and}} House \\
\bottomrule
\end{tabular}
\caption{Top 75 Activating `Extractable' and `Unextractable' Chunks for \texttt{GPT2} on the Subspace Learned at the Transformer Block Output of Layer 6.  The relevant token is highlighted.}
\label{tab:chunk-gpt2}
\end{table*}
\end{document}